\documentclass[%
aip,
amsmath,amssymb,
reprint,%
]{revtex4-1}

\usepackage{graphicx}
\usepackage{dcolumn}
\usepackage{bm}

\usepackage[utf8]{inputenc}
\usepackage[T1]{fontenc}
\usepackage{mathptmx}
\usepackage{etoolbox}

\usepackage{caption}
\usepackage{subcaption}
\usepackage{makecell}
\usepackage{comment}

\usepackage{xcolor}
\def\red{\color{black}}

\newcommand{\VSpike}{V_{\mathrm{Spike}}}
\newcommand{\OUTWeight}{V_{\mathrm{OUT(Weight)}}}
\newcommand{\VRing}{V_{\mathrm{Ring}}}
\newcommand{\VWeight}{V_{\mathrm{Weight}}}
\newcommand{\VSYN}{V_{\mathrm{SYN}}}
\newcommand{\VIN}{V_{\mathrm{IN}}}
\newcommand{\CSYN}{C_{\mathrm{SYN}}}
\newcommand{\Cmem}{C_{\mathrm{mem}}}
\newcommand{\Vmem}{V_{\mathrm{mem}}}
\newcommand{\VFire}{V_{\mathrm{Fire}}}
\newcommand{\Vth}{V_{\mathrm{th(Fire)}}}

\makeatletter
\def\@email#1#2{%
	\endgroup
	\patchcmd{\titleblock@produce}
	{\frontmatter@RRAPformat}
	{\frontmatter@RRAPformat{\produce@RRAP{*#1\href{mailto:#2}{#2}}}\frontmatter@RRAPformat}
	{}{}
}%

\makeatother
\begin{document}
	
	\preprint{AIP/123-QED}
	
	\title{CMOS-based area-and-power-efficient neuron and synapse circuits for time-domain analog spiking neural networks}
	\author{Xiangyu Chen}
	\thanks{X. Chen and Z. Byambadorj contributed equally to this work.}%
	\affiliation{ 
		Systems Design Lab., School of Engineering, the University of Tokyo, Tokyo, Japan
	}
	\author{Zolboo~Byambadorj}
	\thanks{X. Chen and Z. Byambadorj contributed equally to this work.}%
	\affiliation{ 
		Systems Design Lab., School of Engineering, the University of Tokyo, Tokyo, Japan
	}
	
	\author{Takeaki Yajima}%
	\affiliation{ 
		Department of Electrical and Electronic Engineering, Kyushu University, Fukuoka, Japan
	}%
	
	\author{Hisashi Inoue}
	\affiliation{ 
		National Institute of Advanced Industrial Science and Technology (AIST), Ibaraki, Japan
	}%
	\author{Isao H. Inoue}
	\affiliation{ 
		National Institute of Advanced Industrial Science and Technology (AIST), Ibaraki, Japan
	}%
	\author{Tetsuya~Iizuka}
	\homepage{iizuka@vdec.u-tokyo.ac.jp}
	\affiliation{ 
		Systems Design Lab., School of Engineering, the University of Tokyo, Tokyo, Japan
	}%
	
	\date{\today}
	
	\begin{abstract}
		Conventional neural structures tend to communicate through analog quantities such as currents or voltages, however, as CMOS devices shrink and supply voltages decrease, the dynamic range of voltage/current-domain analog circuits becomes narrower, the available margin becomes smaller, and noise immunity decreases. More than that, the use of operational amplifiers (op-amps) and continuous-time or clocked comparators in conventional designs leads to high energy consumption and large chip area, which would be detrimental to building spiking neural networks. In view of this, we propose a neural structure for generating and transmitting time-domain signals, including a neuron module, a synapse module, and two weight modules. The proposed neural structure is driven by a leakage current of MOS transistors and uses an inverter-based comparator to realize a firing function, thus providing higher energy and area efficiency compared to conventional designs. The proposed neural structure is fabricated using TSMC~65~nm CMOS technology. The proposed neuron and synapse occupy the area of 127~$\bm{\mu m^2}$ and 231~$\bm{\mu m^2}$, respectively, while achieving millisecond time constants.
		Actual chip measurements show that the proposed structure implements the temporal signal communication function with millisecond time constants, which is a critical step toward hardware reservoir computing for human-computer interaction. 
	Simulation results of the spiking-neural network for reservoir computing with the behavioral model of the proposed neural structure demonstrate the learning function. 
	\end{abstract}
	
	\maketitle
	
	Deep neural networks (DNNs), which are the second generation of artificial neural networks (ANNs), have extensively explored in recent years for growing number of applications. However, their huge energy consumption especially for the memory access in conventional von-Neumann architecture has forced people to find an alternative way to achieve more power-efficient solutions~\cite{Nature_Zhang,IEEE_Shin,Nature_LeCun,AIP_Kohno,APL_Chicca,JAP_Bo}. Spiking neural network (SNN) is one of the attractive solutions as the third generation of ANNs that can realize learning function with low power by mimicking biological neurons. SNNs consist of neurons and synapses, and are usually built using a bottom-up approach, which means that each component of the SNNs needs to be designed first\cite{JAP_Bo,APL_Yang,JJAP_Chen,Neural_Networks_Maass,NC_Radhakrishnan,SSDM_Chen,JAP_Jeong}. 
	
	
	Many hardware implementations of pulsed neurons or synapses have been reported\cite{TNW_Indiveri,TCASII_Wu,IJCNN_Joubert,TBCAS_Aamir,TCASI_Basu,Sci_Dutta,TCASI_Rubino,TCASI_Aamir,SSCI_Moradi}. To implement the leaky integrate function of neurons, conventional designs usually build integrators with operational amplifiers (op-amps)\cite{TCASII_Wu} and often use large on-chip capacitors and resistors to mimic the millisecond time constants of biological neurons\cite{TCASI_Basu,TBCAS_Aamir}. Moreover, to implement the neuron ``fire'' function, a dedicated circuit structure of a continuous-time or clocked comparator is usually used to set the threshold for neuron excitation\cite{TCASI_Aamir,TCASII_Wu,IJCNN_Joubert,TNW_Indiveri,TBCAS_Aamir}. The bias current of the continuous-time comparator undoubtedly increases the power consumption of the neuron, while the clocked comparator requires additional clock signal distribution and the complex comparator structure occupies a large chip area. While more advanced processes can achieve low power consumption by reducing supply voltage and static leakage current\cite{SSCI_Moradi}, this also leads to a narrower dynamic range, smaller available margin, and degraded noise immunity of the voltage/current-domain analog circuits\cite{IEICE_Asada}. This is detrimental to conventional neural networks that use analog quantities such as voltage and current to communicate with each other. 
	On the other hand, thanks to the scaled transistors that have an improved operation speed with sharp signal transitions, the analog information can be represented more efficiently in time domain, i.e. a time interval of two signal transitions. This so-called time-domain circuit have another advantage in its power efficiency as it often consists of inverters or logic gates that ideally consume no DC power~\cite{JSSC_Staszewski,IEICE_Asada}. Thus, time-domain circuits are ideal for future implementations of low-power SNNs.  
	
	In this paper we propose an original neural structure for generating and transmitting time-domain signals to compose a time-domain neural network. The integrated structure includes neuron and synapse modules that respectively generate and transmit time-domain signals, as well as weight modules for learning functions. 
    One of our main target applications is reservoir computing, which processes information related to human activity. Our application targets simpler and less data-intensive processing such as bio-signals. 
	In reservoir computing, learning functions such as ECG and speaker recognition as well as handwriting recognition can be implemented using only a few hundreds of neurons. Ref.~\onlinecite{Gallicchio_Neural} shows that learning performance improves when the time constants of the input effects are matched between the target function and reservoir dynamics, and we use millisecond time constants as a design target for a neural structure that will be used to process time-series information of human activities. We use the behavioral model of the proposed neural structure to construct the SNN for reservoir computing and implement the learning function, which proves that our proposed neural structure can be used for reservoir computing.

	\begin{figure}[t]
		\centering
		\begin{subfigure}[t]{0.28\textwidth}
			\centering
			\includegraphics[width=\textwidth]{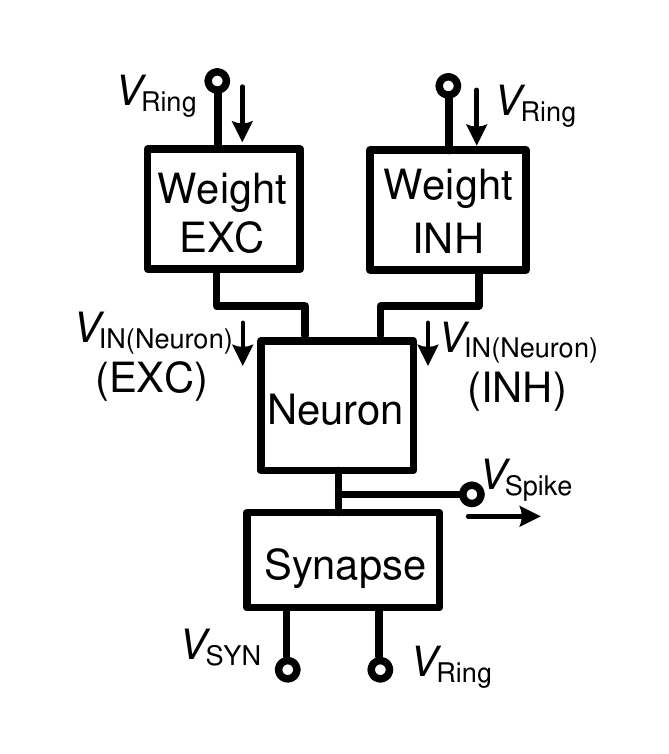}
			\caption{}
			\label{Architecture1}
		\end{subfigure}
		\hfill
		\begin{subfigure}[t]{0.18\textwidth}
			\centering
			\includegraphics[width=\textwidth]{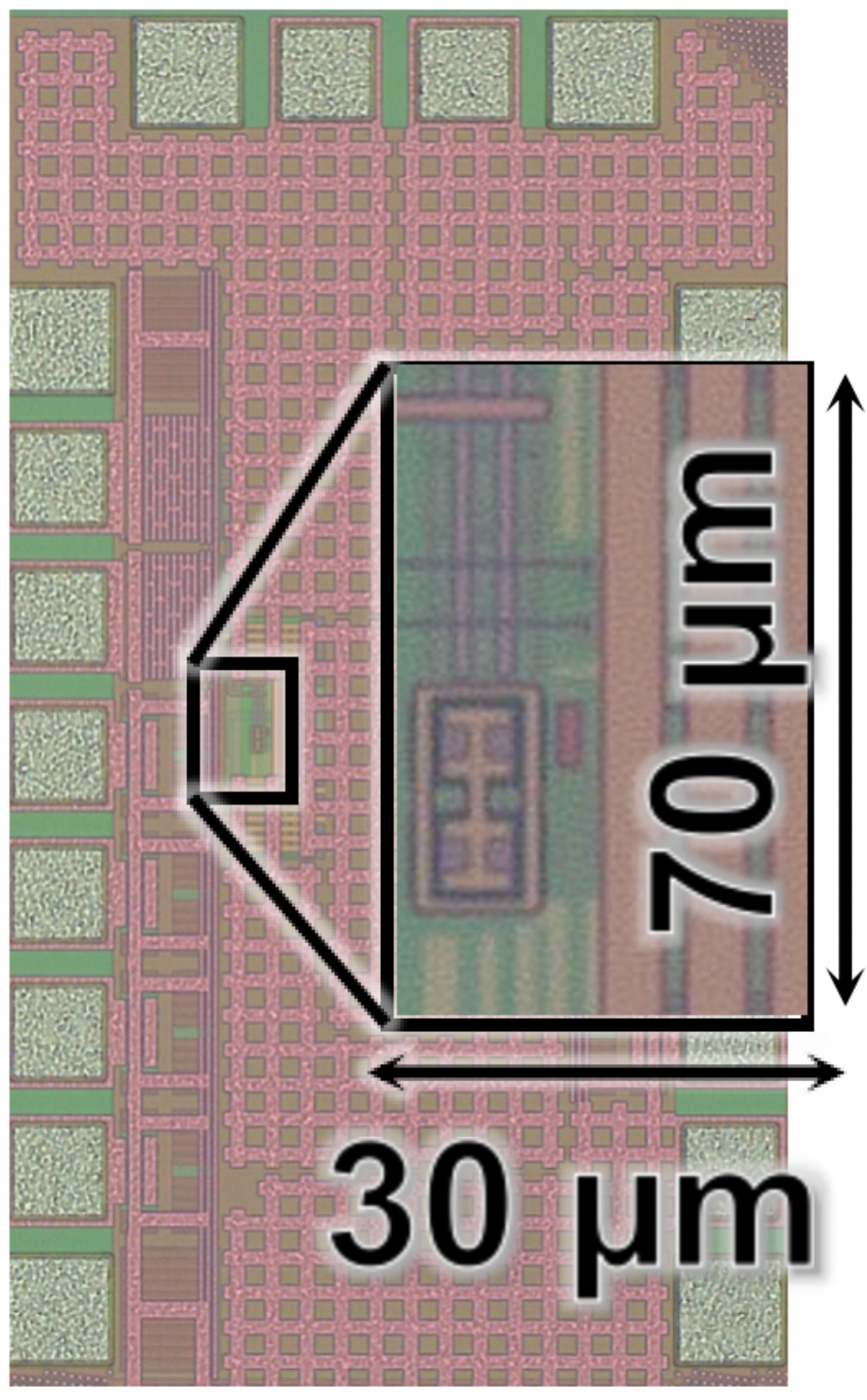}
			\caption{}
			\label{chipphoto}
		\end{subfigure}
		\caption{(a) The proposed structure and (b) a micrograph of the chip.}
		\label{Proposed_Arch}
	\end{figure}
	The designed and fabricated neural structure is shown in Fig.~\ref{Architecture1}, which is based on the proposed neuron, synapse, and weight modules, which will be described in detail below. In this structure, the input of the neuron module is connected to two weight modules, one for tuning the inhibitory signal and the other for the excitatory signal. 
	We fabricated the proposed neural structure shown in Fig.~\ref{Architecture1} with TSMC 65~nm standard CMOS technology. 
	The micrograph of the chip is shown in Fig.~\ref{chipphoto}, where the die area of neuron, synapse and weight modules are 127~$\rm{\mu m^2}$, 231~$\mu m^2$ and 525~$\mu m^2$, respectively. 
	
	The LIF neuron model consists mainly of a membrane capacitor, a leaky resistor, and a voltage comparator. Neurons receive signals from other neurons via synapses and the soma generates action potentials in response to these external signals. If a neuron receives a sufficient number of spikes through the synapse, its membrane potential will reach a threshold value, causing the neuron to ``fire''\cite{Book_Dayan,Book_Gerstner,JJAP_Chen}.
	
	\begin{figure*}[tbp]
		\centering
		\begin{subfigure}[t]{1.1\columnwidth}
			\centering
			\includegraphics[width=\columnwidth]{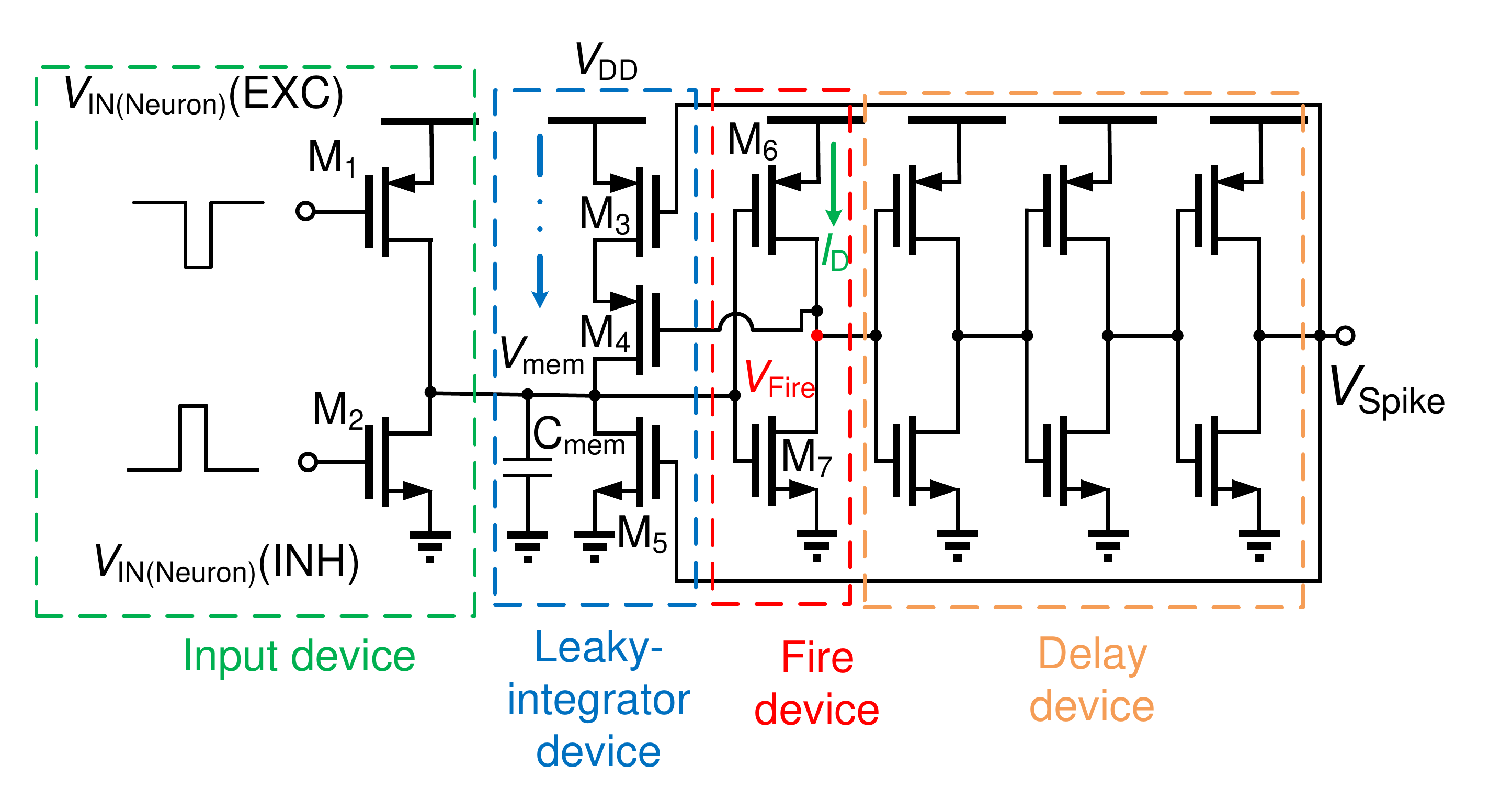}
			\caption{}
			\label{Proposed_Neuron}
		\end{subfigure}
		\begin{subfigure}[t]{0.8\columnwidth}
			\centering
			\includegraphics[width=\columnwidth]{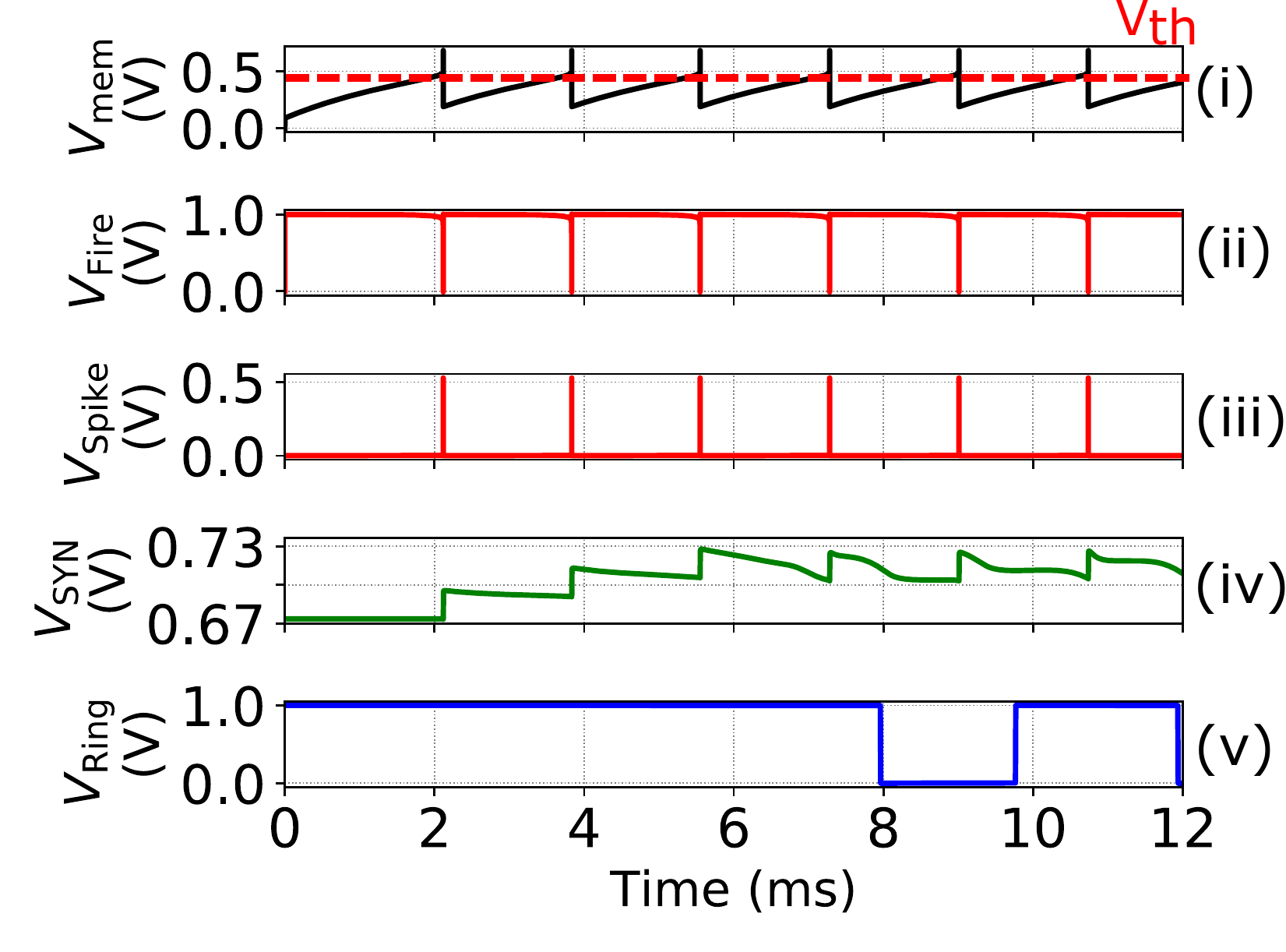}
			\caption{}
			\label{Beh_Proposed_LIF}
		\end{subfigure}
		\begin{subfigure}[t]{1\columnwidth}
			\centering
			\includegraphics[width=\columnwidth]{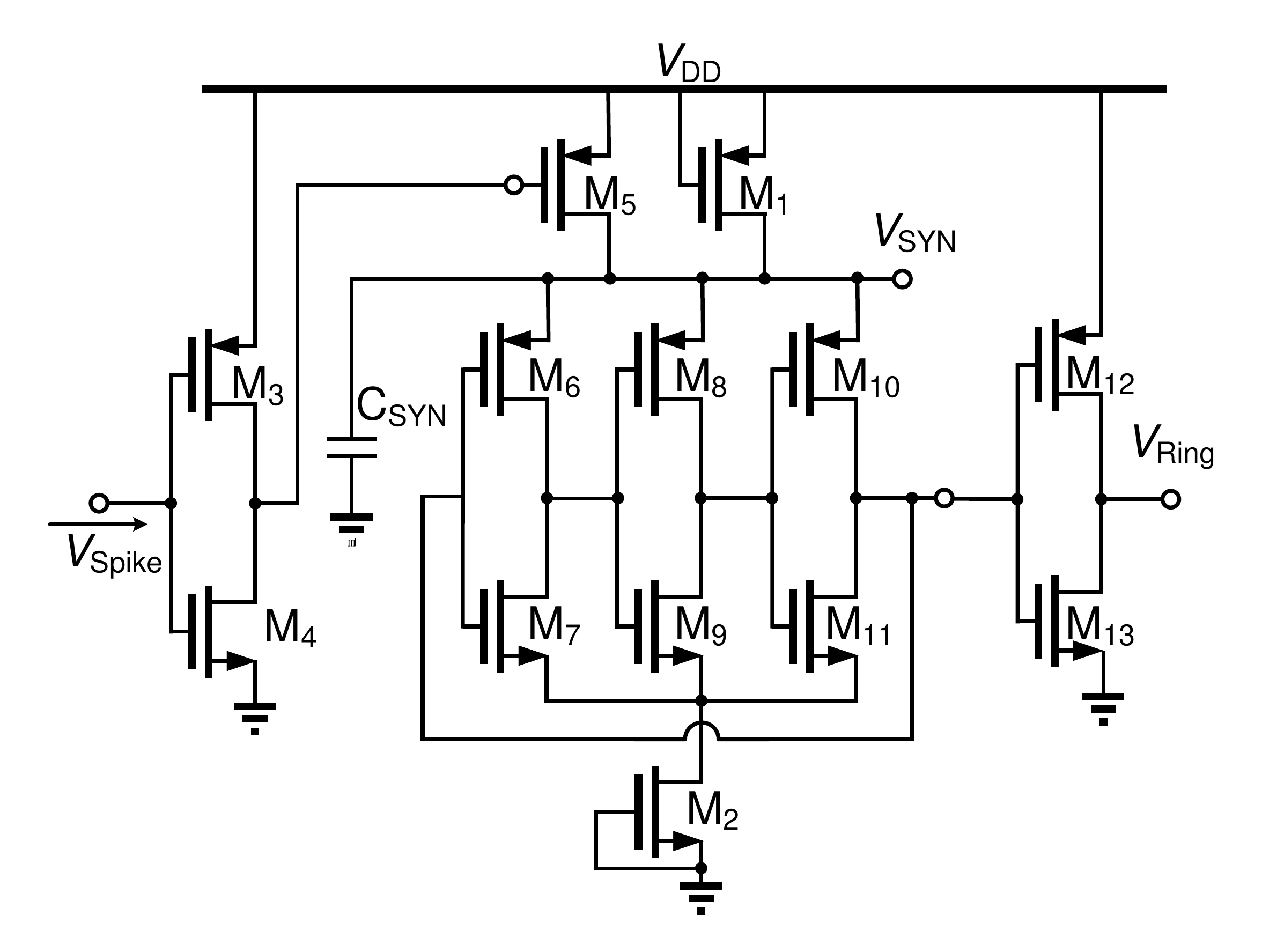}
			\caption{}
			\label{Proposed_Synapse}
		\end{subfigure}
		\begin{subfigure}[t]{1.05\columnwidth}
			\centering
			\includegraphics[width=\columnwidth]{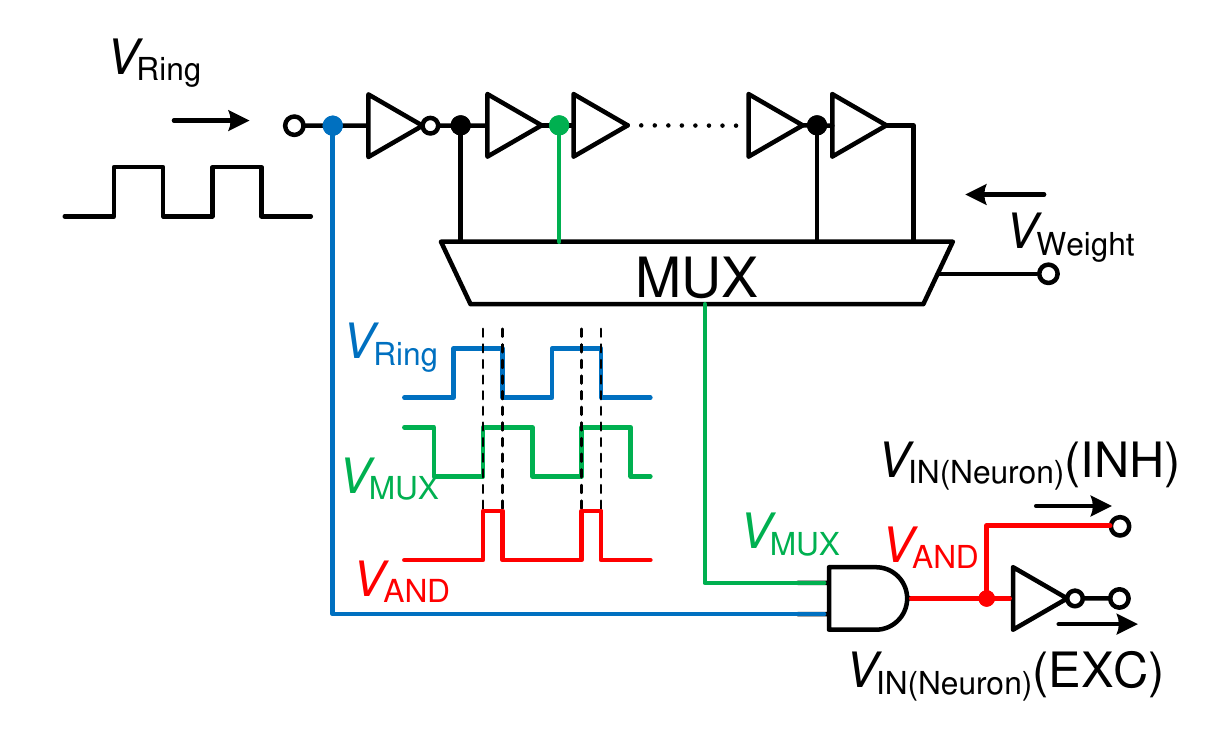}
			\caption{}
			\label{Proposed_weight}
		\end{subfigure}
		\caption{(a)~Circuit diagram of the proposed neuron module, (b)~behaviours of proposed LIF neuron and synapse modules, (c)~circuit diagram of the proposed synapse module, and (d)~circuit diagram of the proposed weight module.}
	\end{figure*}
	
	The use of inverters to implement the ``fire'' function is already known as an alternative to comparators. Ref.~\onlinecite{Sci_Yajima} has proposed an inverter-based neuron, which is well suited for use in the proposed neural structure, 
	and therefore the neuron used in this study was designed based on Ref.~\onlinecite{Sci_Yajima}, which is shown in Fig.~\ref{Proposed_Neuron}. It consists of input device, leaky integrator device, fire device and delay device.  Originally in Ref.~\onlinecite{Sci_Yajima}, the circuit is not assumed to be designed as an element to build a neural network, thus does not have a structure to receive excitatory and inhibitory signals. In the proposed circuit, on the other hand, the input device that consists of $\mathrm{M}_\mathrm{1}$ and $\mathrm{M}_\mathrm{2}$ receives an excitatory input and an inhibitory input, respectively. The inputs to $\mathrm{M}_\mathrm{1}$ and $\mathrm{M}_\mathrm{2}$ are narrow pulse signals as shown in Fig.~\ref{Proposed_Neuron}, which is generated from a pre-stage synapse. The activity of the pre-stage synapse is represented by the pulse frequency, and the coupling weight is represented by the pulse width.
	When more than one pre-stage synapses are connected to compose a network, the multiple pulses can be applied through OR logic, or by adding input devices connected in parallel. With the parallel input devices, the neuron circuit can accept multiple pulses even at the same time. 

  In leaky-integrator device, $\Cmem$ represents the cell membrane of the neuron, and $\mathrm{M}_\mathrm{5}$ can be regarded as a leaky resistor in the resting state. When there is no external input to the input device, the capacitor is charged by the leakage current of $\mathrm{M}_\mathrm{3}$ and $\mathrm{M}_\mathrm{4}$,  and the membrane potential $\Vmem$ increases  continuously with the inflow of the leakage current (the current is integrated as shown in Fig.~\ref{Beh_Proposed_LIF}(i)). At this point, since $\mathrm{M}_\mathrm{5}$ is in off state, it can be considered as a resistor in parallel with the capacitor, i.e. leaky resistor, capable of achieving a long time constant. 
	
	Once $\Vmem$ rises to the threshold voltage $\Vth$, the fire device is activated (Fig.~\ref{Beh_Proposed_LIF}(ii)). In conventional designs, LIF neurons mostly use dedicated circuit structures of continuous-time or clocked comparator to set the threshold voltage. This is not friendly for building SNNs that are as energy efficient and bio-scale as the brain. In this study, the fire device is implemented by an inverter-based comparator that can set the threshold voltage by two transistors instead of a continuous-time or clocked comparator. 
    {\red To realize an accurate threshold voltage for an inverter-based comparator, we may use an auto-zeroing technique that periodically sense, store and cancel the offset with switches and capacitors~\cite{DataConv_Razavi}. However, it requires multiple-phase clocks to control the switches, thus is not suitable for area-and-power-efficient reservoir implementations.}
	Though {\red with a simple inverter-based comparator} there may be a threshold variation due to process, voltage and temperature fluctuations, it can be seen as mimicking the difference between individuals of real neurons. In addition, the learning function is able to compensate for threshold differences and process variations~\cite{Front_Wunderlich}.  
	When there is an excitatory pulse input, $\mathrm{M}_\mathrm{1}$ will be turned on instantaneously, which causes more current to charge $\Cmem$ and $\Vmem$ to rise rapidly. Conversely, an inhibitory pulse input signal will cause $\mathrm{M}_\mathrm{2}$ to turn on momentarily, causing $\Cmem$ to charge slower or even discharge through $\mathrm{M}_\mathrm{2}$, which in turn slows down the rate of $\Vmem$ rise or makes it fall. 
	
	\begin{figure*}[tbp]
		\centering
		\begin{subfigure}[b]{0.45\textwidth}
			\centering
			\includegraphics[width=\columnwidth]{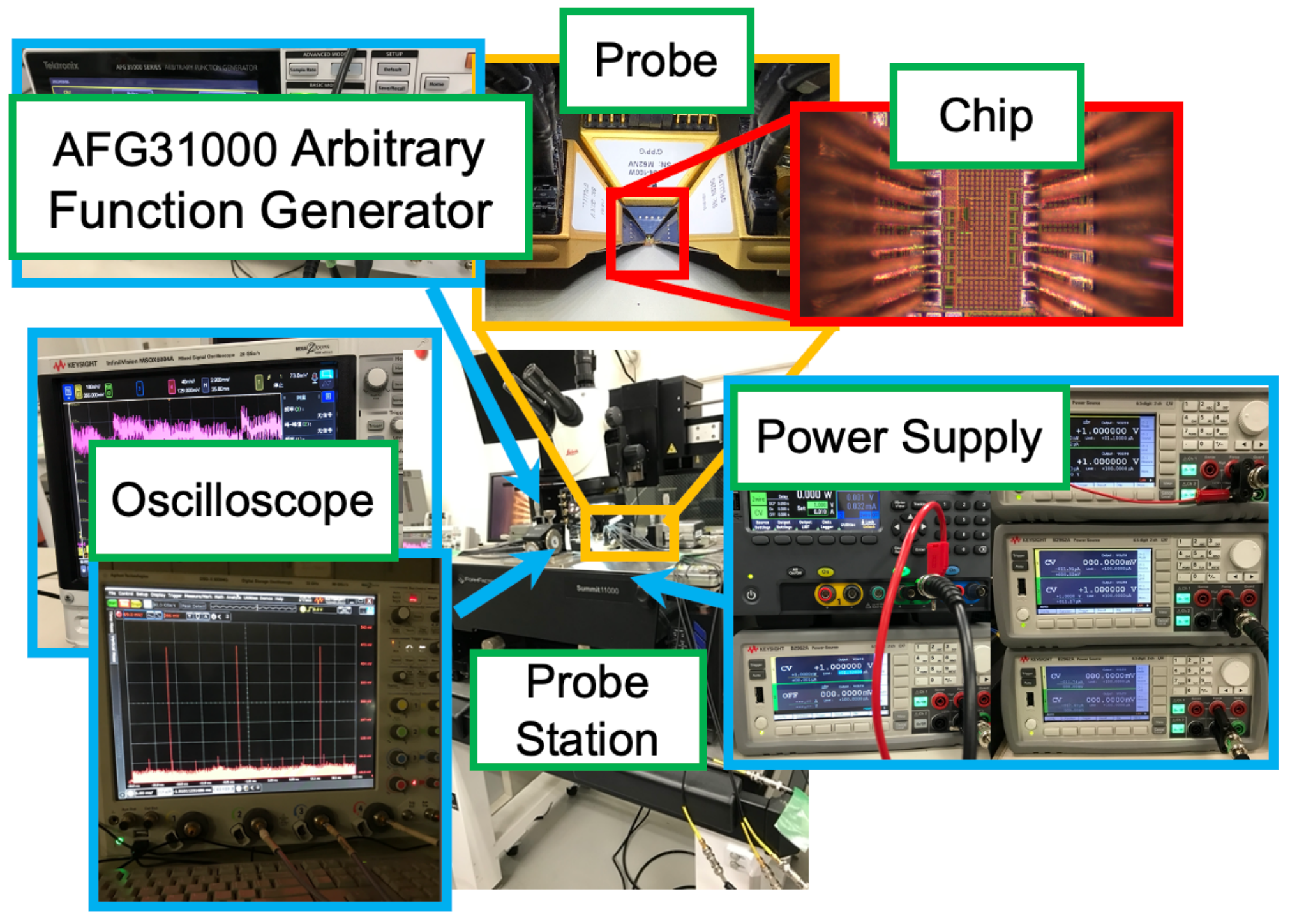}
			\caption{}
			\label{experimental setup}
		\end{subfigure}
		\hfill
		\begin{subfigure}[t]{1\columnwidth}
			\centering
			\includegraphics[width=\columnwidth]{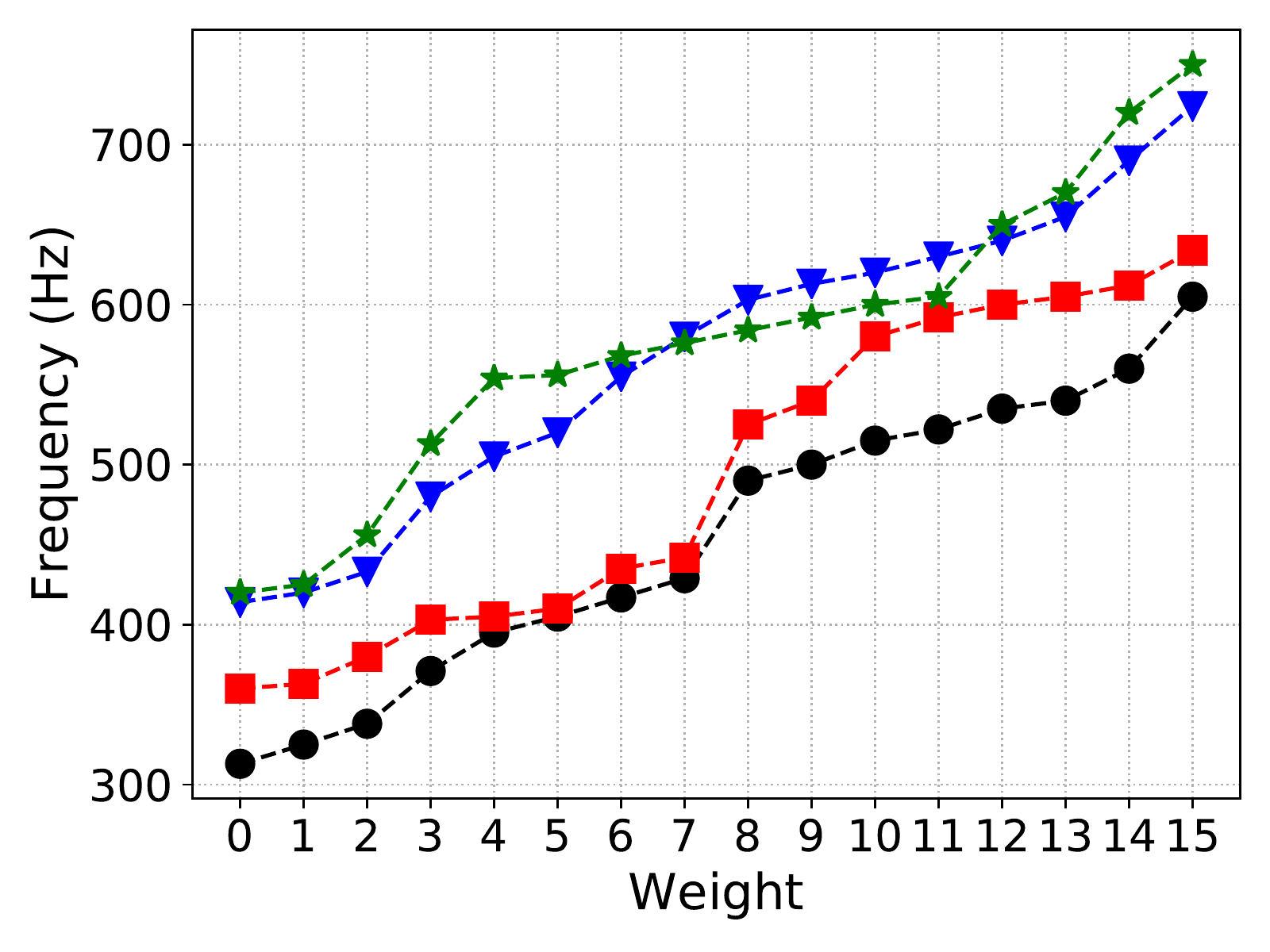}
			\caption{}
			\label{weight_effect}
		\end{subfigure}
		\hfill
		\begin{subfigure}[t]{1\columnwidth}
			\centering
			\includegraphics[width=\columnwidth]{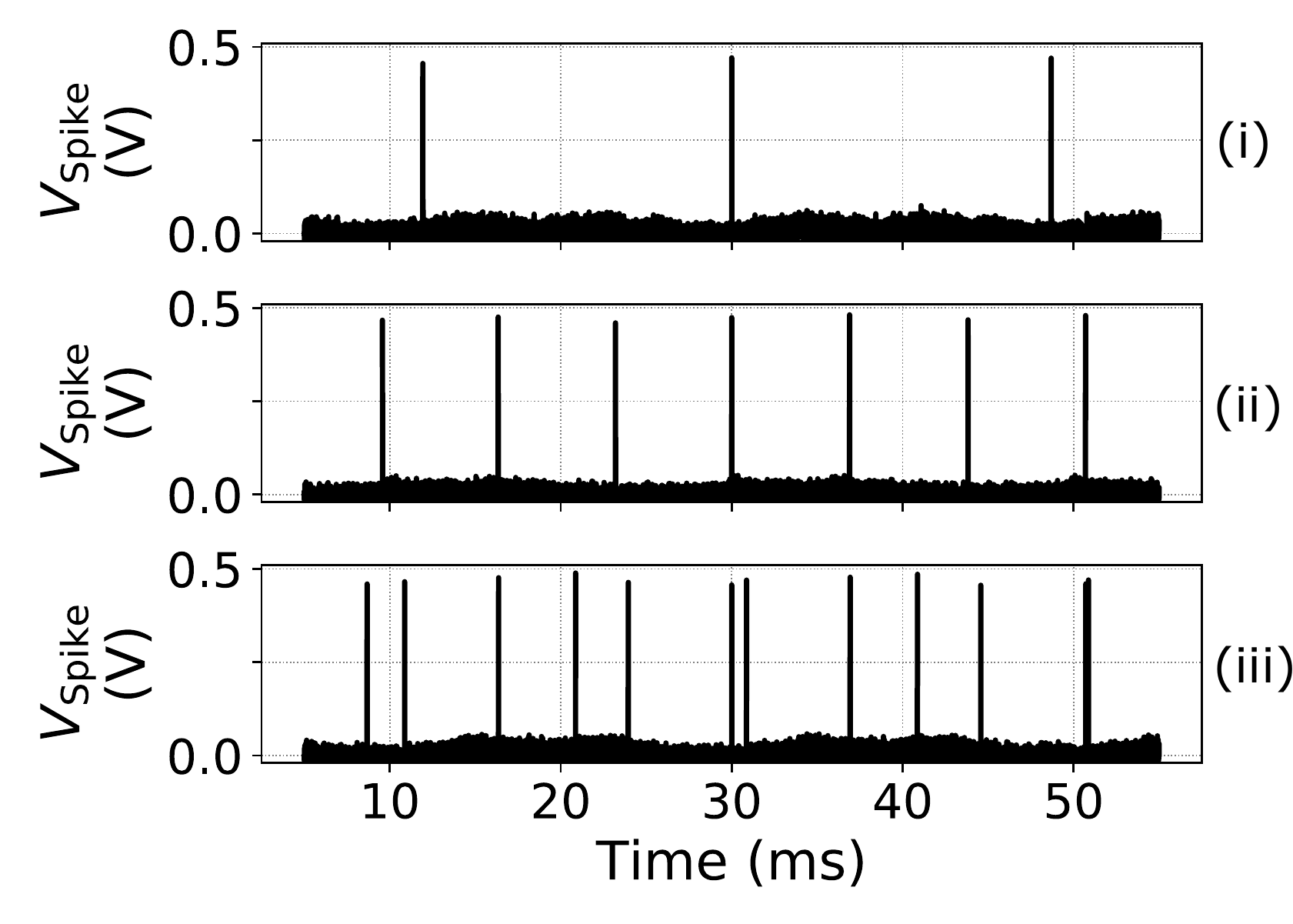}
			\caption{}
			\label{Spike}
		\end{subfigure}
		\hfill
		\begin{subfigure}[t]{1\columnwidth}
			\centering
			\includegraphics[width=\columnwidth]{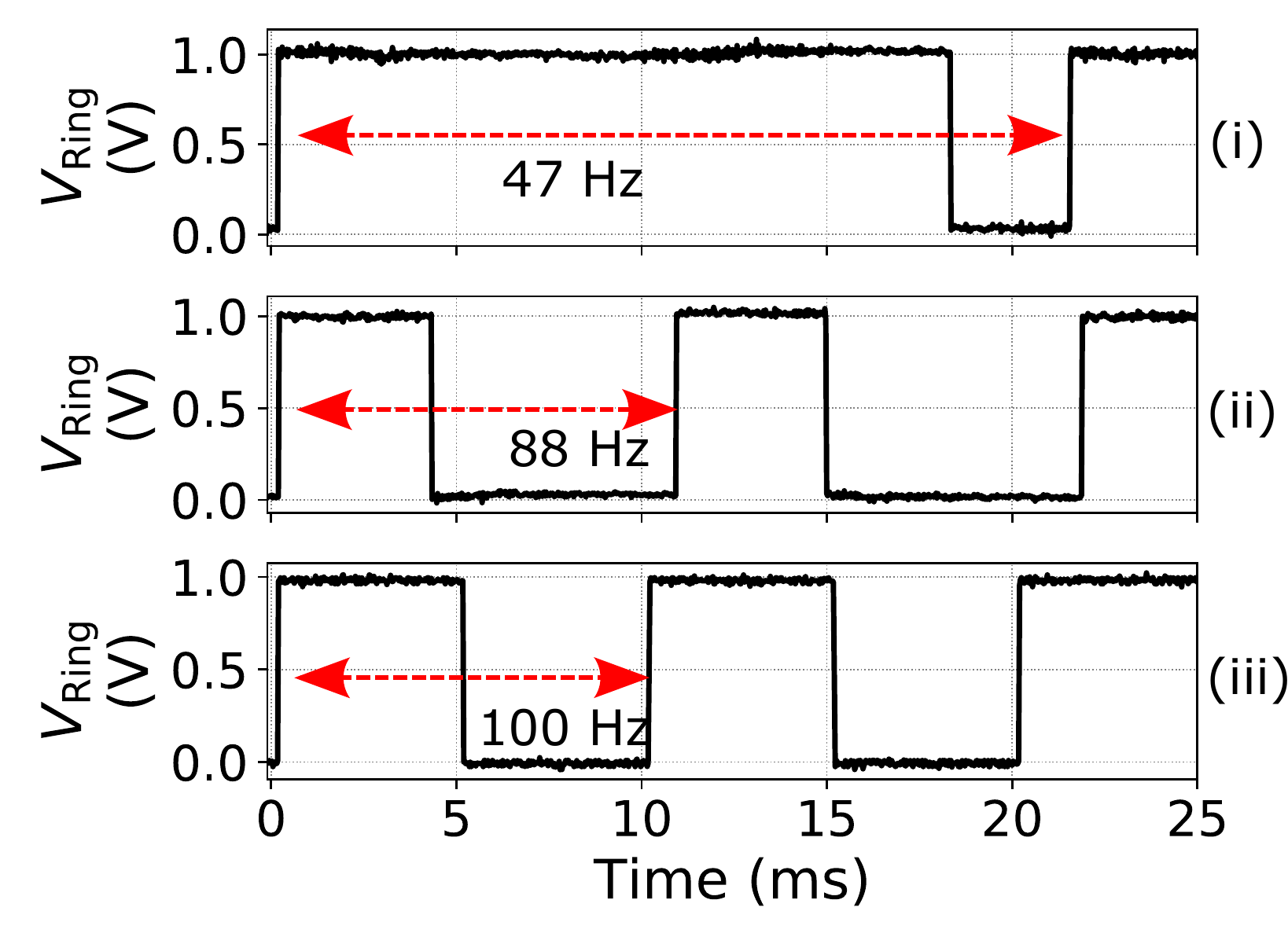}
			\caption{}
			\label{Ring}
		\end{subfigure}
		\caption{(a)~A photo of the experimental setup, (b)~the measured firing rate of the neuron for 4 chips, (c)~the measured waveforms of the neuron output, and (d)~the measured waveforms of the synapse output.}
	\end{figure*}
	
	\begin{figure*}[t]
		\centering
		\begin{subfigure}[t]{0.23\textwidth}
			\centering
			\includegraphics[width=\textwidth]{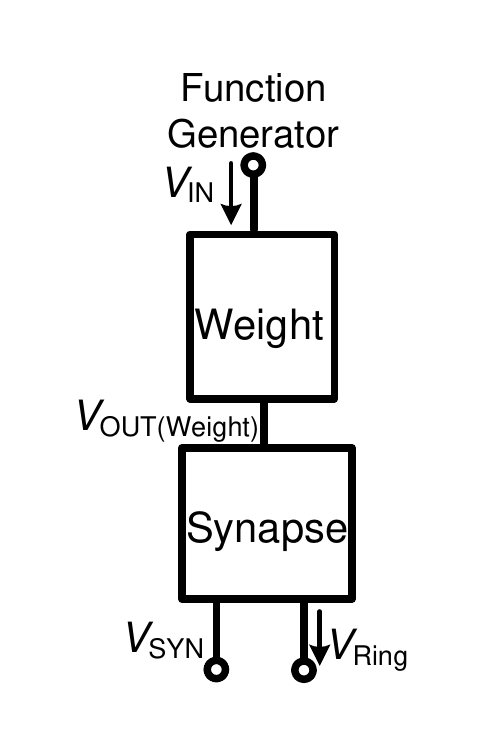}
			\caption{}
			\label{synchronous1}
		\end{subfigure}
		\hspace{0.08\columnwidth} 
		\begin{subfigure}[t]{0.45\textwidth}
			\centering
			\includegraphics[width=\textwidth]{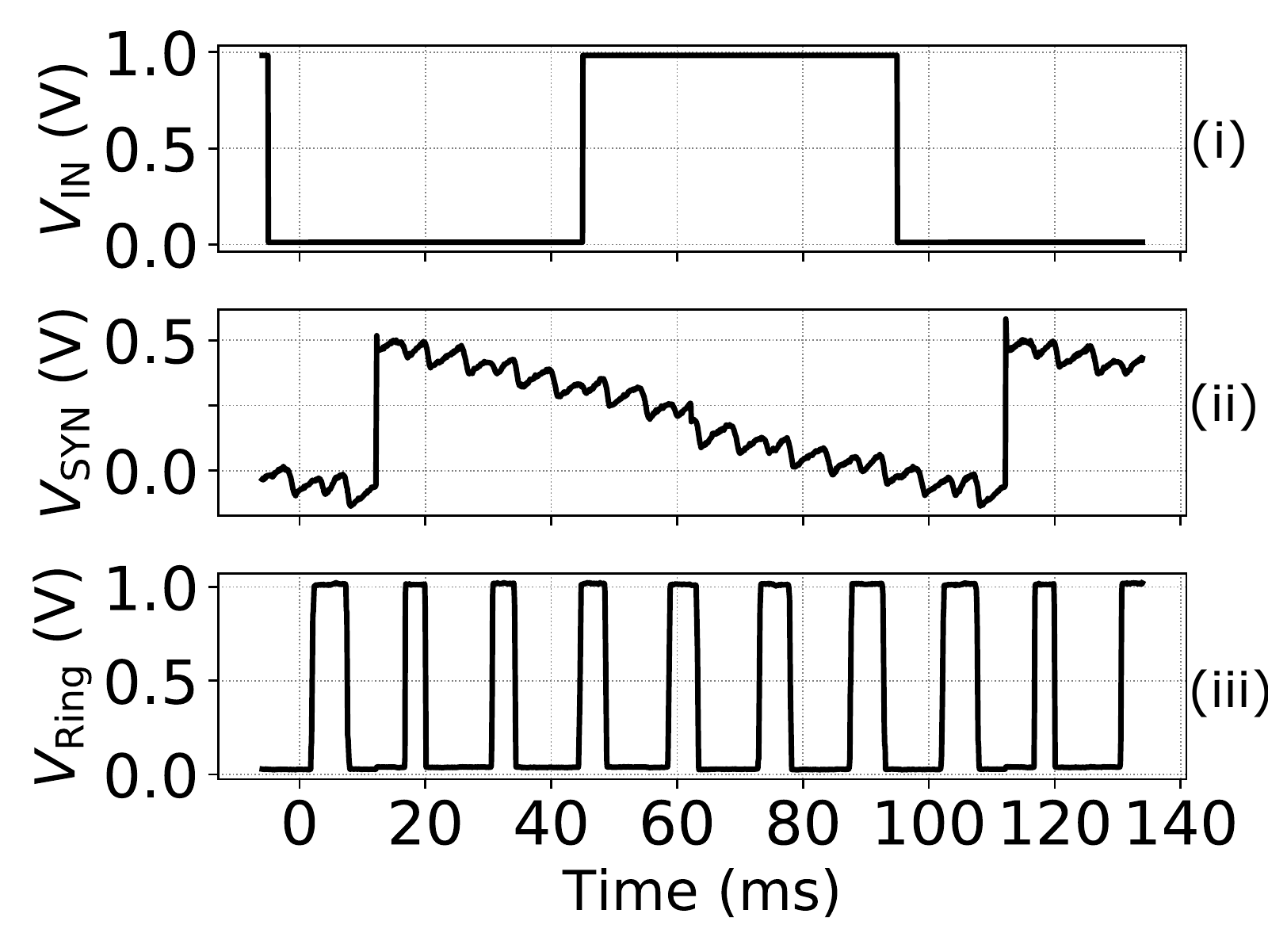}
			\caption{}
			\label{synchronous2}
		\end{subfigure}
		\caption{(a) Another combined structure fabricated to evaluate the synapse and (b) the measured waveforms of $\VRing$ and $\VSYN$.}
		\label{Proposed_Archi2}
	\end{figure*}
	
	\begin{figure*}[t]
		\centering
		\begin{subfigure}[c]{0.44\textwidth}
			\centering
			\includegraphics[width=\textwidth]{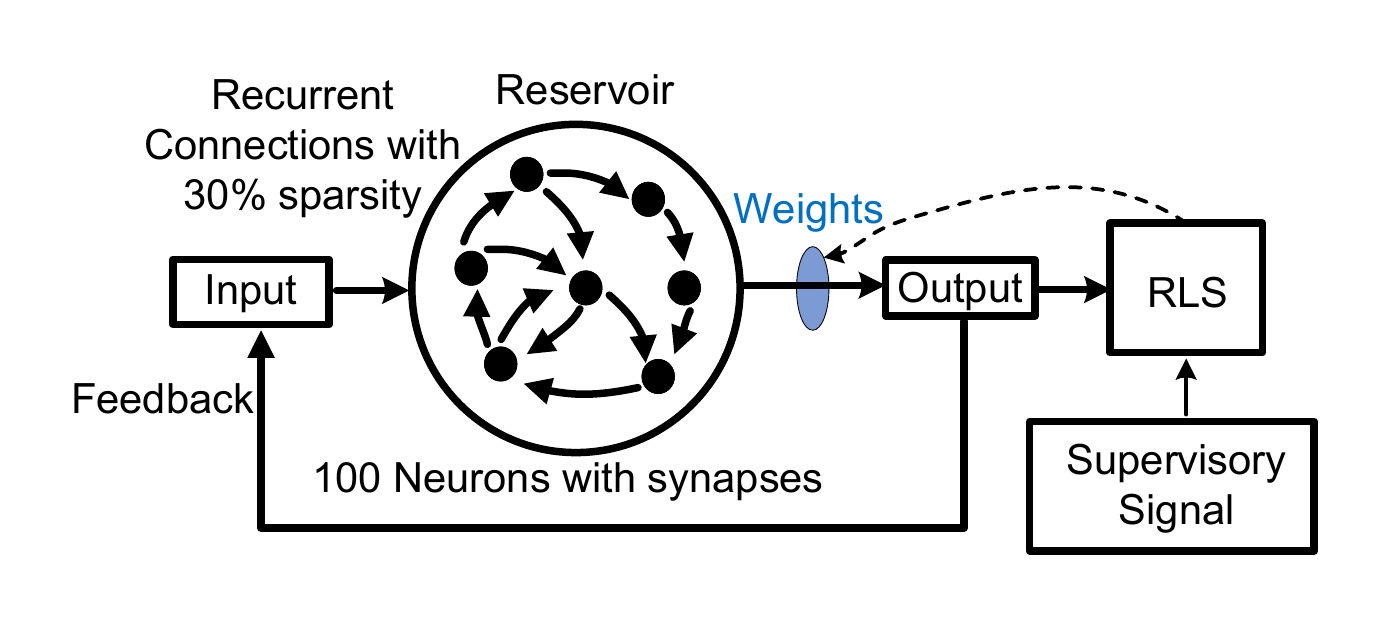}
			\caption{}
			\label{simulation_model}
		\end{subfigure}
			\hfill
		\hspace{0.0\columnwidth} 
		\begin{subfigure}[c]{0.54\textwidth}
			\centering
			\includegraphics[width=\textwidth]{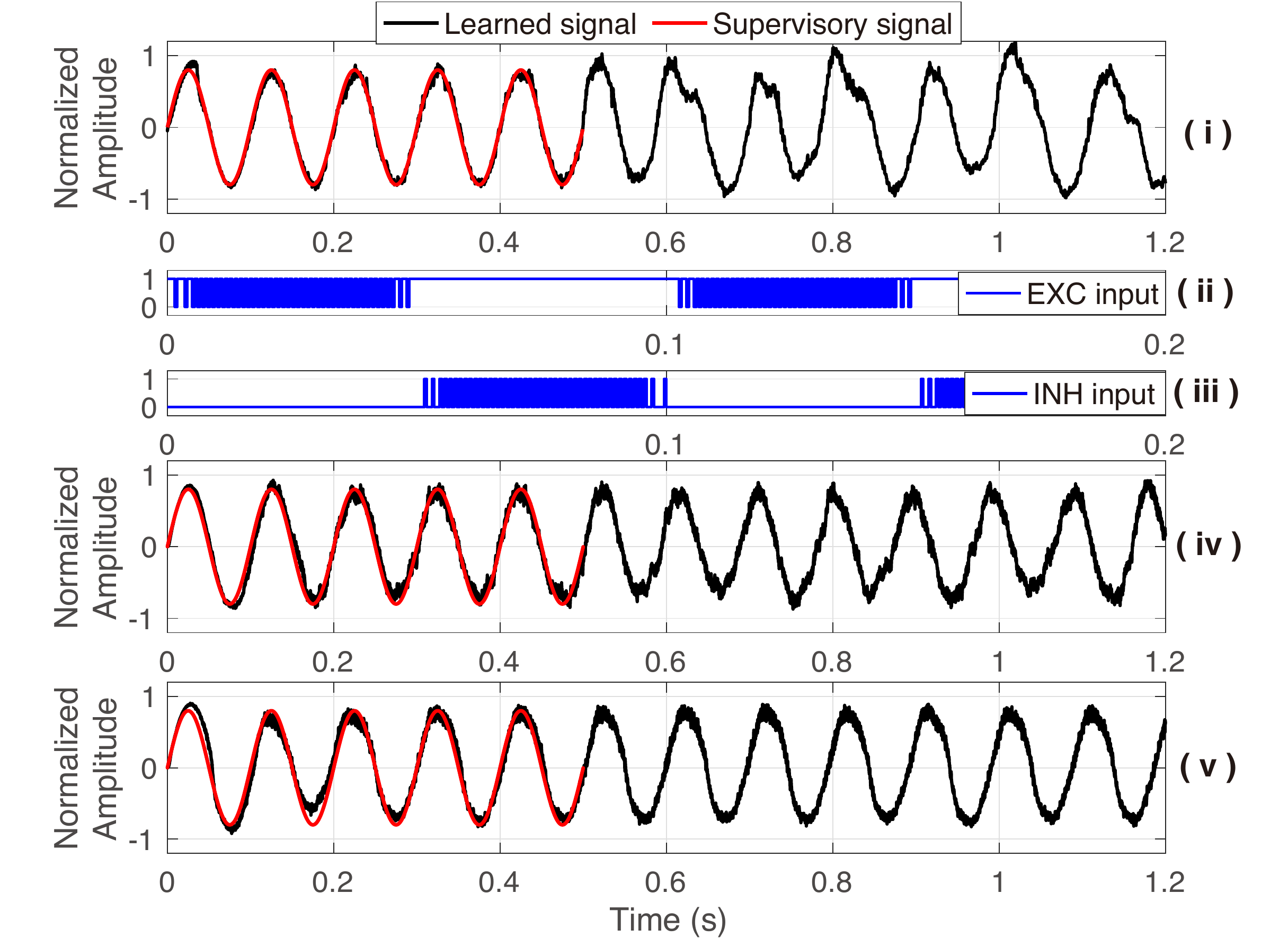}
			\caption{}
			\label{SNN_Learning}
		\end{subfigure}
    \caption{(a) The behavioral model of the SNN for reservoir computing based on the proposed neural structure. (b) The system-level behavioral simulation results: (i) based on a model with $15$\,Hz$ - 200$\,Hz frequency tuning range, zoomed-in view of the (ii) excitatory and (iii) inhibitory input signals converted from the output, {\red (iv)} based on $15$\,Hz$ - 2$\,kHz and {\red(v)} $15$\,Hz$ - 20$\,kHz frequency tuning ranges.}
		\label{behavioral}
	\end{figure*}

	When the fire device is activated, it generates a low level of $\VFire$ to be connected to $\mathrm{M}_\mathrm{4}$, which will increase the current to charge the membrane capacitor $\Cmem$, resulting in an instantaneous increase of the membrane potential $\Vmem$, which promotes the triggering of the fire device. This mimics the influx of Na$^+$ into the cell membrane prompting a rapid increase in membrane voltage, i.e., a positive feedback effect. 
	Finally, the low level of $\VFire$ generated from the fire device is converted to a high level of $\VSpike$ (Fig.~\ref{Beh_Proposed_LIF}(iii)) by a delay device that includes a three-stage inverter and connects the $\VSpike$ to $\mathrm{M}_\mathrm{3}$ and $\mathrm{M}_\mathrm{5}$, resetting $\Vmem$ to zero. This process mimics the activation of K$^+$ channels in biological neurons, resulting in the outward flow of K$^+$ ions and the eventual return of the cell membrane to its resting state. 

	Synapses are essential modules in SNNs, as neurons are interconnected by them. We have designed a neuron module for generating time-domain signals, and then we need a transmission medium, i.e., a synapse, to transmit this time-domain signal to other neurons. 
	To compose a complete neural network, we design a synapse module based on frequency signals, as shown in Fig.~\ref{Proposed_Synapse}. The synapse consists mainly of a voltage-controlled ring oscillator operating under a leakage current, which is composed of a three-stage inverter ($\mathrm{M}_\mathrm{6}, \mathrm{M}_\mathrm{7}, \mathrm{M}_\mathrm{8}, \mathrm{M}_\mathrm{9}, \mathrm{M}_\mathrm{10}$ and $\mathrm{M}_\mathrm{11}$). 
	The previous neuron circuit fires and generates a spike $\VSpike$, which is inverted by an inverter, making $\mathrm{M}_\mathrm{5}$ open for a short time, and the current flowing through $\mathrm{M}_\mathrm{5}$ charges $\CSYN$, which will increase $\VSYN$. Once $\VSYN$ reaches the voltage that triggers the oscillation, the ring oscillator begins to oscillate (Fig.~\ref{Beh_Proposed_LIF}(iv) and Fig.~\ref{Beh_Proposed_LIF}(v)). 
	If the preceding neuron does not fire for a long time, $\VSYN$ will leak until the initial state, at which point the synapse becomes inactive again. Since $\VSYN$ is equivalent to the supply voltage of the ring oscillator, the current flowing out of $\mathrm{M}_\mathrm{5}$ controls $\VSYN$ and thus the frequency of the ring oscillator. 
	
	SNNs achieve learning function by adjusting the weights; therefore, we propose a weight module that is compatible with the proposed time-domain neuron and synapse modules described above, as shown in Fig.~\ref{Proposed_weight}. The proposed weight module tunes the time-domain information, which is the width of the output pulses. This module consists of a delay line, a multiplexer, and an AND gate. $\VRing$ is the square wave signal from the synapse that will pass through the delay line. $\VWeight$ is the digital code that represents weight, which is determined after learning and is used to control the multiplexer.
	The width of the output pulse that corresponds to the time-domain weight is adjusted according to which tap in the inverter chain is selected by the multiplexer. As mentioned earlier, if the excitatory or inhibitory pulse width is wide, the voltage $\Vmem$ in the subsequent neuron is charged or discharged faster, respectively. This corresponds to a large weight. In this study, we chose a multiplexer with 16 inputs, i.e. four bit weights (0000 to 1111). 
	The output of the weight module is connected to the input device of the subsequent neuron circuits. The frequency of the pulse (pulse spacing) and the width of the pulse act simultaneously on the neuron to change its activity. The frequency of the pulse is determined by the output frequency of the previous synapse, while the coupling strength depends on the width of the pulse output determined by the weight module. 
	
	Figure ~\ref{experimental setup} shows the experimental setup used to test the fabricated neural structure chip (Fig.~\ref{chipphoto}), where the chip was placed on a probe station Summit11000 and tested with probes in direct contact with it. 
	In the experiments, we assume that the inputs of the two weight modules is the pre-stage synapses, which is emulated by the arbitrary function generators.
	The output of the neuron is connected to the synapse module, and the output of which will be varied in response to the change in the output of the neuron. 
	We used a Tektronix AFG31252 arbitrary function generator as a pre-stage synapse to provide square wave signals for our fabricated neural circuits. At the same time, we observed the output waveforms using oscilloscopes (Keysight MSOX6004A and DSOX93304Q). 
	
	The experimental results are shown in Figs.~\ref{weight_effect},~\ref{Spike}~and~\ref{Ring}. 
	To verify the effect of weights on the firing rate of neurons, we fixed the frequency of the pre-stage synapse output (function generator) at 100 Hz and observed the change in the firing rate of neurons for 4 chips by adjusting the weights module. We averaged the spike frequencies 1024 times over a time range of 100 ms to derive the corresponding neuronal firing frequency under each weight setting, as shown in Fig.~\ref{weight_effect}. The proposed neuron is basically firing with the rate determined by the leakage currents into and out from $\Cmem$ in balance, and input from the previous stage modulates it. We can see that when the weights become larger, the neuron module firing frequency becomes larger. Mainly due to the process variation of the FETs, the firing frequency fluctuates in about $\pm$10--17\,\% over 4 chips.
    {\red Especially for the use in a reservoir, however, due to the random weights in its recurrent connections, these random variations should be compensated for during the learning process in the output weights.}
	
	Figure~\ref{Spike} compares the variation of neuron fire times depending on the signal from the pre-stage synapse. The insets (i), (ii)~and~(iii) of Fig.~\ref{Spike} show the cases with 100 Hz inhibitory input (weight is set to 1100), no input, and with 100 Hz excitatory input (weight is set to 1100), 	respectively, from which we can see that the inhibitory input decreases the fire frequency of the neuron and increases the fire interval, while the excitatory input works as the opposite of the inhibitory input. The experimental results show that the firing interval of the proposed neuron is on the order of milliseconds, which is in accordance with the feature of biological neurons having millisecond time constants. 
	When no signal is fed from the pre-stage synapse, the power consumption is about 800~pW, generating about 20 spikes in a 100~ms cycle. From this, it can be roughly estimated that each spike consumes about 4~pJ of energy.
	Subsequently, the insets~(i),~(ii)~and~(iii) of Fig.~\ref{Spike} were used as input signals to the synapse to influence $\VRing$. 
	The measured $\VRing$ waveforms in these three cases are shown in Fig.~\ref{Ring}. The average of the frequencies for each case measured in 5~s time period are 41~Hz, 90~Hz and~98~Hz, respectively. The feasibility of this synapse output frequency range will be validated with system-level simulations in the following discussion. 
	
	To facilitate the observation of the synchronous response of the synapse, we also fabricated the structure of Fig.~\ref{synchronous1}.  Figure~\ref{synchronous2} is the experimental results on Fig.~\ref{synchronous1}. 
	We used a Tektronix AFG31252 arbitrary function generator to generate a 10~Hz square wave signal $\VIN$ as shown in Fig.~\ref{synchronous1}~(i). After $\VIN$ passes through a weight module, it produces a spike signal $\OUTWeight$. 
	The voltage $\VSYN$ is observed through an on-chip source follower as an analog buffer. 
    Though $\OUTWeight$ is not designed to be observed from outside as it is a narrow pulse, with the arrival of the $\OUTWeight$ after the falling edge of $\VIN$, the $\VSYN$ voltage at the synapse rises instantaneously as shown in Fig.~\ref{synchronous2}(ii), which in turn increases the frequency of $\VRing$. If the $\OUTWeight$ does not arrive for a long time, $\VSYN$ decreases, which in turn affects the $\VRing$ frequency to become smaller. 
	
	\begin{table}[t]
		\caption{\label{Comparison}Performance Comparison of Stand-Alone Neuron Circuits.}
		\begin{ruledtabular}
			\begin{tabular}{ccccccc}
				Ref. & Technology & \makecell{Energy per \\ spike (pJ)} & \makecell{Single\\neuron\\area \\ ($\mu m^2$)} & \makecell{Frequency\\(Hz)} & \makecell{Neuron \\model} & \makecell{Sim. or \\Meas.}\\
				\hline
				\onlinecite{TNW_Indiveri}&  \makecell{350 nm \\ CMOS} & 900 & 2573 &100
				& IF & Meas. \\
				\onlinecite{TCASII_Wu}& \makecell{180 nm \\ CMOS} & 9.3 & $10^4$ &N/A
				& LIF & Meas.\\
				\onlinecite{IJCNN_Joubert}& \makecell{65 nm \\ CMOS} & 41 & 538 & 300
				& LIF & Sim.\\
				\onlinecite{TBCAS_Aamir}&  \makecell{65 nm \\ CMOS} & 200 & 3363 &$1.9\times 10^{6}$
				& AdEx-IF\footnotemark[1] & Meas.\\
				\onlinecite{Sci_Dutta}& \makecell{32 nm \\ SOI \\MOSFET} & 35 & 1.8 &$10^{6}$ 
				& LIF & Meas.\\
				\onlinecite{TCASI_Rubino}& \makecell{22 nm \\ FD-SOI} & 14 & 900 & 30
				&  AdEx-IF\footnotemark[1] & Meas.\\
				\onlinecite{SSCI_Moradi}& \makecell{22 nm \\ CMOS} & 1.17/0.36 & 70 & 30/100
				&  LIF & Sim.\\
				\makecell{this \\work}& \makecell{65 nm \\ CMOS} & 4 & 127 & 230
				& LIF & Meas.\\
			\end{tabular}
		\end{ruledtabular}
		\footnotetext[1]{Adaptive-exponential integrate-and-fire.}
		\vspace{-0.2cm}
	\end{table}
	
	Table \ref{Comparison} shows the performance comparison among stand-alone neuron circuits. The proposed neuron circuit has advantages in terms of energy consumption and area. The designs in Refs.~\onlinecite{TNW_Indiveri,TCASII_Wu,IJCNN_Joubert,TBCAS_Aamir} used a continuous-time or a clocked comparator, and these designs take up a large amount of chip area as well as power consumption. 
	The neuron fabricated in a non-CMOS process proposed in Ref.~\onlinecite{Sci_Dutta} does not require a comparator, which leads to an advantage in area. However, its energy consumption is relatively high and these particular technologies are less mature and thus more costly compared to standard CMOS processes. 
    Both Ref.~\onlinecite{TCASI_Rubino} and Ref.~\onlinecite{SSCI_Moradi} are being fabricated in an advanced process. However, compared to this work  Ref.~\onlinecite{TCASI_Rubino} does not have an advantage in terms of energy consumption and area. Although Ref.~\onlinecite{SSCI_Moradi} shows better energy efficiency with simulation results, when normalized by the technology node, the proposed neuron achieves better area efficiency.

To demonstrate the feasibility of the proposed spiking neuron and the ring oscillator-based synapse circuits, a behavioral simulation is carried out in MATLAB environment as shown in Fig.~\ref{simulation_model}. In this simulation, 100 neurons are used with random recurrent connections with the proposed synapse and weighting modules.  
The proposed weight modules are applied only in the reservoir layer and their weights are assigned randomly in advance and fixed during the learning process. 
{\red Thus the random fluctuations in the reservoir 
are compensated for during the learning process in the output weights.} 
To establish a realistic simulation, the output frequency range of each synapse is set from $15$\,Hz to $200$\,Hz based on the actual measurement results.
The recursive least square (RLS) algorithm is used to train the output weights as introduced in Ref.~\onlinecite{SUSSILLO2009544}. A $10$\,Hz sinewave, which corresponds to the time-scale of human-activity-related information, is used as an example of supervisory input signal. The supervisory and the trained output signal are shown in Fig.~\ref{SNN_Learning}(i). The feedback signal from the output is converted to excitatory and inhibitory pulse trains whose frequencies are in proportion to the absolute value of the output amplitude as shown in Figs.~\ref{SNN_Learning}(ii) and (iii), respectively. After 5 periods of supervisory signal, the output weights are fixed and the SNN generates the learned signal by itself, which demonstrates the feasibility of the proposed neural structures for learning function. 
We have also found from these simulations that to further improve the learning capability the output frequency tuning range of the synapse should be increased, which can be done by optimizing the synapse circuit. For example, with the extended frequency tuning ranges from $15$\,Hz$ - 2$\,kHz and $15$\,Hz$ - 20$\,kHz, the learned signals become smoother to better reproduce the supervisory signal as shown in {\red Figs.~\ref{SNN_Learning}(iv) and \ref{SNN_Learning}(v)}, respectively. 


In summary, we have proposed a neural structure for generating and transmitting time-domain signals. The proposed neuron and synapse occupy an area of 127~${\mu m^2}$ and 231~${\mu m^2}$, respectively. This structure does not use op-amps and continuous-time or clocked comparators, while the firing function is realized with an inverter-based comparator to provide advantages in area and power consumption. The proposed time-domain neural structure benefits from scaled process technologies compared to conventional voltage/current-domain designs. Actual chip fabrication and measurement results demonstrate the temporal signal communication function with millisecond time constants.
The proposed time-domain neural structure is well suited for building spiking neural networks for processing real-time time-series information for human-computer interaction.
\\

This work is supported by Japan Science and Technology Agency (JST) CREST Grant Number JPMJCR19K2.
\nocite{*}
\bibliography{APL_Chen}

\end{document}